\documentclass{article}

\usepackage{PRIMEarxiv}

\usepackage[utf8]{inputenc} 
\usepackage[T1]{fontenc}    
\usepackage{hyperref}       
\usepackage{url}            
\usepackage{booktabs}       
\usepackage{amsfonts}       
\usepackage{nicefrac}       
\usepackage{microtype}      
\usepackage{lipsum}
\usepackage{fancyhdr}       
\usepackage{graphicx}       
\graphicspath{{media/}}     

\usepackage{amsmath}
\usepackage{algorithmic}
\usepackage{algorithm}
\usepackage{array}
\usepackage[caption=false,font=normalsize,labelfont=sf,textfont=sf]{subfig}
\usepackage{textcomp}
\usepackage{stfloats}
\usepackage{xcolor}
\usepackage{verbatim}
\usepackage{graphicx}
\usepackage{adjustbox}
\usepackage{booktabs}
\usepackage{cite}
\usepackage{times}
\usepackage{epsfig}
\usepackage{subfig}
\usepackage{comment}

\graphicspath{{figures/}}     
\DeclareGraphicsExtensions{.pdf,.png,.jpg}

\title{ADRMX: Additive Disentanglement of Domain Features with Remix Loss
}

\author{
  Berker Demirel \\
  Sabanci University \\
  Istanbul \\
  \texttt{\{berkerdemirel\}@sabanciuniv.edu} \\
   \And
  Erchan Aptoula \\
  Sabanci University \\
  Istanbul\\
  \texttt{erchan.aptoula@sabanciuniv.edu} \\
  \And
  Huseyin Ozkan \\
  Sabanci University \\
  Istanbul \\
  \texttt{huseyin.ozkan@sabanciuniv.edu} \\
}

\begin{document}
\maketitle

\begin{abstract}
The common assumption that train and test sets follow similar distributions is often violated in deployment settings. Given multiple source domains, domain generalization aims to create robust models capable of generalizing to new unseen domains. To this end, most of existing studies focus on extracting domain invariant features across the available source domains in order to mitigate the effects of inter-domain distributional changes. However, this approach may limit the model's generalization capacity by relying solely on finding common features among the source domains. It overlooks the potential presence of domain-specific characteristics that could be prevalent in a subset of domains, potentially containing valuable information. In this work, a novel architecture named Additive Disentanglement of Domain Features with Remix Loss (ADRMX) is presented, which addresses this limitation by incorporating domain variant features together with the domain invariant ones using an original additive disentanglement strategy. Moreover, a new data augmentation technique is introduced to further support the generalization capacity of ADRMX, where samples from different domains are mixed within the latent space. Through extensive experiments conducted on DomainBed under fair conditions, ADRMX is shown to achieve state-of-the-art performance. Code will be made available at GitHub after the revision process.
\end{abstract}


\section{Introduction}
Over the past decade, deep learning systems have achieved remarkable success across different tasks. However, their performance is often evaluated under the assumption that train and test data follow the same, or similar distributions \cite{he2016deep, krizhevsky2017imagenet, simonyan2014very, szegedy2015going}. In real-world scenarios the assumption that train and test data are independent and identically distributed is violated due to the changes in background, illumination, occlusion, scale, camera angle and other factors. These distributional change between train and test set are commonly referred to as domain shift \cite{pan2009survey}. Addressing the domain shift problem has become a significant focus of research, as conventional deep neural network architectures tend to learn and adapt to the specific statistical properties of the training data, which may not be present in the test set \cite{nam2021reducing, yan2020improve, li2018learning}. Consequently, these models usually fail to generalize well to unseen domains. To tackle this challenge, numerous studies are performed within two different scenarios: domain adaptation \cite{ganin2015unsupervised, bousmalis2016domain}, and domain generalization \cite{muandet2013domain, li2017deeper}. Unlike domain adaptation, domain generalization models do not assume access to the target domain during training. Therefore, the objective is to extract essential and transferable knowledge from the source domains, enabling effective generalization to unseen target domains. This makes domain generalization particularly more challenging, as the models should learn to capture the underlying essence of the data and generalize to different data statistics.

\begin{figure}[t]
  \begin{center}
    \subfloat[]{\label{fig:fig1_a}\includegraphics[width=0.225\textwidth]{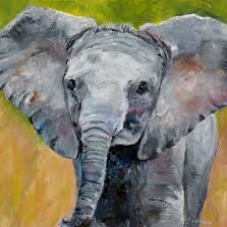}}
    \hspace{0.2\textwidth}
    \subfloat[]{\label{fig:fig1_b}\,\includegraphics[width=0.225\textwidth]{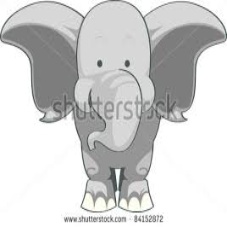}}\\
    \subfloat[]{\label{fig:fig1_c}\includegraphics[width=0.225\textwidth]{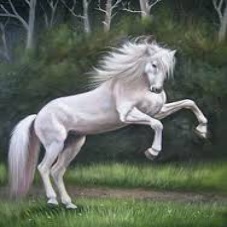}}
    \hspace{0.2\textwidth}
    \,\subfloat[]{\label{fig:fig1_d}\includegraphics[width=0.225\textwidth]{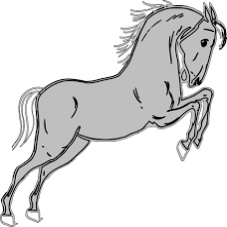}}
  \end{center}
  
  \caption{Example images from the PACS dataset \cite{li2017deeper} showcasing the persistence of domain specific attributes despite significant domain shifts. The first row displays images of the elephant class, while the second row features images of the horse class from the art and cartoon domains, respectively.}
  \label{fig:fig1}
\end{figure}

In this paper, a novel approach called Additive Disentanglement of Domain Features with Remix Loss (ADRMX) is presented, that tackles the domain shift problem under a domain generalization scenario. The disentanglement of domain variant and domain invariant features is achieved using an additive approach, enabling the model to capture contextual characteristics of objects. Moreover, the additive modeling allows domain-specific features from different samples to be effectively blended with domain-invariant features, thus enriching the data in the latent space. This not only populates the training data but also ensures the network adapts to diverse distributions by incorporating instances from multiple sources. We hypothesize that, in contrast to previous works \cite{ganin2015unsupervised, li2018domain, ganin2016domain}, incorporating domain variant features alongside domain invariant features provides an "additional" guide that improves generalization. Figure \ref{fig:fig1} demonstrates that even when domains are significantly different, they can share certain characteristics, and learning domain-specific information from one domain can potentially improve performance on others.

Our technique (Figure \ref{fig:model_figure}), ADRMX, consists of two backbones for label and domain feature extraction. They are meant to fulfill the label classification and domain classification tasks effectively with their corresponding losses. After subtracting domain features from label features, we adopt an adversarial learning setting, where resulting features cannot identify domain characteristics while retaining the label relevant information \cite{ganin2015unsupervised}. The additive modeling of the relationship between label features and domain features allows us to introduce a data augmentation technique, remix strategy. The remix strategy utilizes two instances from different domains but with the same label, by merging the domain features of one instance with domain invariant features of another using a simple addition operation. To facilitate the model's generalization to data with a diverse of distributional characteristics, the same classification head is employed for both the label features and remixed samples. In this way, ADRMX is regularized with augmented data without introducing additional parameters. Our main contributions are as follows.
\begin{itemize}
    \item We propose a novel architecture called ADRMX, which effectively disentangles the relationship between domain features and domain invariant features in an additive manner.
    \item Leveraging the additive relationship, we introduce a data augmentation technique that enables the mixing of instances from different domains within the latent space.
    \item The effectiveness of ADRMX is demonstrated by achieving state-of-the-art performance in the DomainBed \cite{gulrajani2020search} benchmark. Notably, across the seven datasets, ADRMX achieves an impressive average accuracy of \textbf{67.6\%}, surpassing the performance of previous approaches.
\end{itemize}

\section{Related Work}
Domain Generalization (DG) has become a prominent research area focusing on developing models that can generalize to new domains without relying on labeled data from the target domain. In this section, an overview of the state-of-the-art is provided.

\subsection{Distribution Alignment}
Several methods have been proposed to align the features by regularizing statistical properties of different sources \cite{sun2016deep, li2018mmd,  sagawa2019distributionally, krueger2021out}. \cite{sun2016deep} proposed an effective approach to extract features from sources by matching their second-order statistics using a nonlinear transformation. This approach serves as a strong baseline, as provided by \cite{gulrajani2020search} which minimizes both the mean and covariance differences. Similarly, \cite{li2018mmd} employed an architecture that minimizes the maximum mean discrepancy between pairs of any source domains utilizing an RBF kernel. On the other hand, \cite{sagawa2019distributionally} leveraged robustness by aiming to minimize the worst-case training loss over source domains. The optimization gives higher importance to the respective domain when it incurs a higher loss. \cite{krueger2021out} addressed the distributional shift problem by introducing risk extrapolation. This technique penalizes the network for instances that introduce losses that are lower or higher than the mean, allowing fairness interpretations as it equalizes the risk across different groups.

\subsection{Adversarial Learning}
Adversarial training is an intuitive way to extract invariant features from different domains \cite{ganin2016domain, li2018deep, nam2021reducing}. The pioneering work \cite{ganin2015unsupervised} and \cite{ganin2016domain} tried to extract features by making the features discriminative for the label prediction task while simultaneously making them indistinguishable across domains. They introduced a gradient reversal layer for domain classification task which enables joint training of a feature extractor and a domain classifier. \cite{li2018deep} extended this idea assuming that the conditional distribution remains the same across different domains. Their approach incorporated class prior-normalized, and class-conditional domain classification losses to regularize the feature extractor. Explicitly considering the conditional distribution further enhanced the model's ability to generalize to unseen domains. \cite{nam2021reducing} argued that feature extractor's inductive bias can be eliminated by disentangling style and context features. They proposed a style-agnostic network that aims to learn representations robust to domain-specific style variations. By separating the style and context information, the model becomes more resilient to changes in style across domains, leading to improved generalization performance. Lastly, adversarial domain generalization has been recently extended to a covariate-shift-free setting \cite{Seemakurthy2023dg}.

\subsection{Domain Mixup}
Recently, \cite{zhang2017mixup} has gained significant attention due to its effectiveness as an augmentation technique. It improves the generalization properties of the network by training it with convex combinations of pairs of samples and their labels. \cite{yan2020improve} built upon this idea by mixing up pairs of source domains in the context of domain generalization. In that way, the network is trained with the convex combinations of pairs of samples from different domains and labels, enhancing its ability to handle domain shifts. In a different vein, \cite{xu2021fourier} proposed a Fourier-based approach that exploits that the semantic information is preserved in the phase of the Fourier transform across different domains. By applying an amplitude mixup strategy, they interpolated between different styles while preserving the underlying semantic information.

\subsection{Meta Learning}
Domain generalization problem has also been studied in the context of meta learning frameworks \cite{li2018learning, balaji2018metareg}.
\cite{li2018learning} introduced a model-agnostic training procedure to address domain shifts. Their approach involves synthesizing potential test domains during training to calculate the meta objective. This meta objective ensures that the algorithm's steps aim to decrease the synthesized test error, leading to strong generalization performance. Similarly, \cite{balaji2018metareg} proposed a method that models the optimization process where steps for a domain are performed only if they achieve a good performance on the other domains. By doing so, they guaranteed that each optimization step contributes to achieving good cross-domain generalization. Label Encoder

\begin{figure*}[ht]
    \centering
    \subfloat[\label{fig:model_figure_train}]{
        \includegraphics[  width=18.2cm,
          height=7.6cm,
          keepaspectratio]{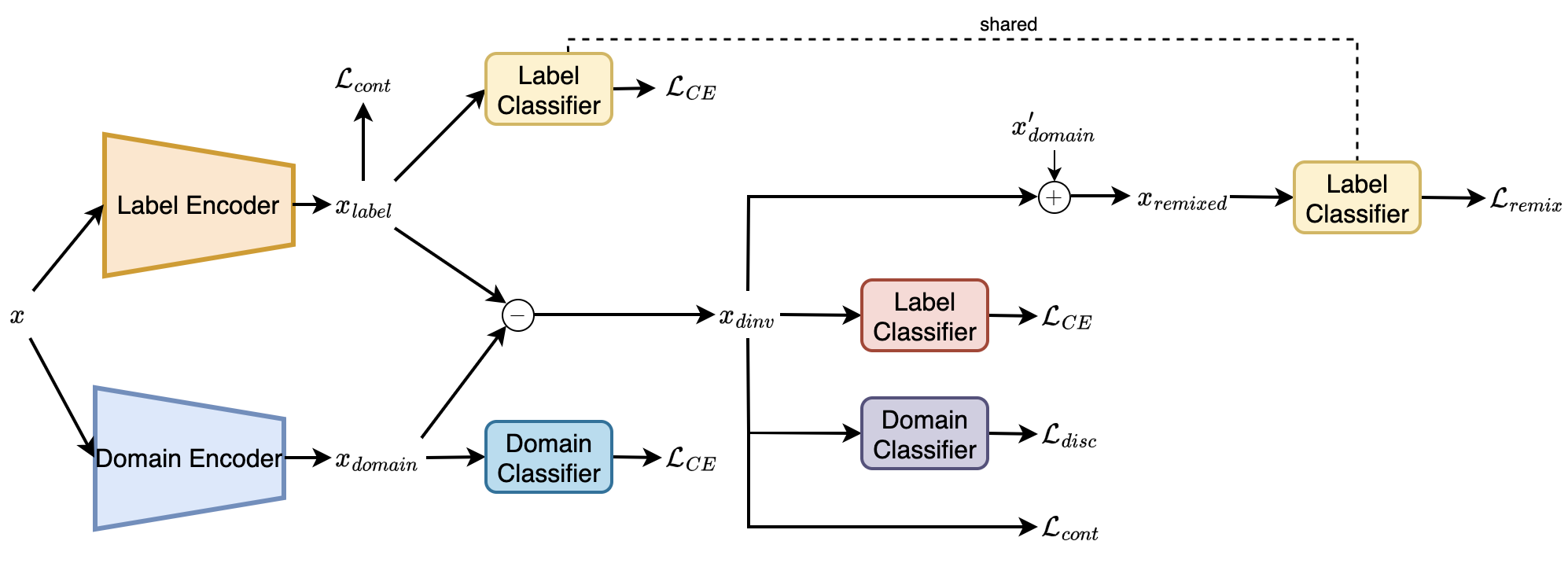}
    }
    \hfill
    \subfloat[\label{fig:model_figure_test}]{
        \includegraphics[width=10.5 cm, height=7 cm, keepaspectratio]{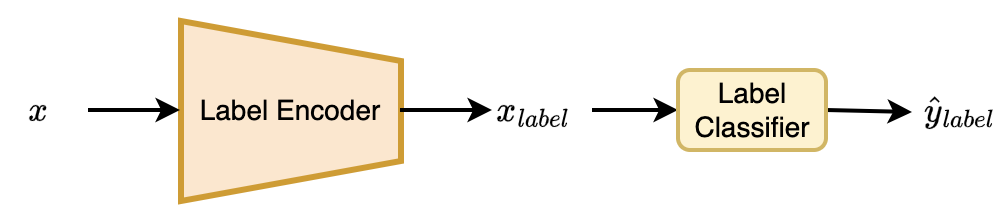}
    }
    \caption{Overview of ADRMX that incorporates domain specific features for prediction. At the training phase (a), label and domain encoders extract their respective features. Domain-invariant features are obtained by subtracting domain features from labels. Cross entropy, contrastive, and domain discrimination losses guide the domain-invariant features to retain label information while discarding domain properties. The remix loss operates on the combined features of the domain-invariant feature of a sample and the domain features of another instance with the same label. During the test phase (b), classification is performed by utilizing label encoder and label classifier.}
    \label{fig:model_figure}
\end{figure*}

\subsection{Contrastive Learning}
Several studies have proposed contrastive learning-based approaches \cite{kim2021selfreg, motiian2017unified}. These methods intuitively facilitate robustness as they attempt to increase the proximity of features belonging to samples with the same class, with respect to a metric. \cite{motiian2017unified} introduced a method to exploit the Siamese architecture along with a contrastive semantic alignment loss, which regularizes the distances between samples from different domains but with the same class label, and samples from different domains and different class labels. Conversely, \cite{kim2021selfreg} highlighted the significance of resolving negative pair sampling to improve generalization performance. They introduced a supervised contrastive learning technique which only uses positive pairs mitigating the challenges emerged from uninformative negative samples.

The proposed method falls into adversarial learning category while utilizing supervised contrastive loss \cite{khosla2020supervised}. Unlike the methods in the prior work, we sought to leverage domain specific features -along with the domain invariant features- that could aid in generalization. Specifically, ADRMX introduces an additive modeling that selectively includes or removes domain information from the feature vector, consisting of two parallel feature extractors for label and domain features. Additive modeling allows many manipulations such as removing and adding other domains, applying orthogonality consistency checks between domain and label features.

\section{ADRMX}
Generalizing to unseen domains is a challenging task which has led the literature to explore methods for extracting domain-invariant features. In that way, by leveraging the mutual information across the source domains, the model can effectively generalize without overfitting the domain-specific features. For instance, in some domains, color distribution may be a significant feature that helps the model capture the label, whereas in other domains, it may not hold the same importance. In that case, an unregularized model would capture an information from that particular domain which is not useful to others. However, it is important to note that not all domain-specific label-relevant features are necessarily worthless. Figure \ref{fig:fig1} contains horse and elephant instances from art and cartoon domains in PACS dataset \cite{li2017deeper}. As they have similar specific color patterns and object compositions, features learned from art paintings can be beneficial for model to recognize objects in cartoons.

To tackle this, we propose a model that is able to handpick the relevant information by disentangling domain variant and domain invariant features in an additive fashion. This additive modeling allows the proposed model to represent label features, domain features and domain invariant features in which domain features are subtracted from label features in order to obtain domain invariant features. Therefore, the model is not limited to using only domain invariant features, but rather can potentially incorporate beneficial domain-specific and label-relevant information. Moreover, the simple element-wise subtraction enables the model to remove domain-specific information from the label, obtaining domain invariant features. These domain invariant features can then be combined with another domain's features by element-wise addition, effectively mimicking its label features.

\subsection{Problem Description}
The problem in domain generalization is to develop a model that can effectively generalize to unseen domains. To evaluate the model's generalization ability, an experimental setup is typically created by training the model on multiple source domains and evaluating its performance on an unseen domain. To assess the model's performance on each domain individually, cross-domain testing is performed, where the model is evaluated on each domain separately, and the average performance across all domains determines the model's success. Therefore, more formally, given the source domains $\mathcal{D} = \{\mathcal{D}_1, \mathcal{D}_2, \dots , \mathcal{D}_S \}$, and the target domain $\mathcal{D}_T$; the objective of domain generalization algorithm is to learn a model using the source domains $\mathcal{D}$, that will perform well on the target domain $\mathcal{D}_T$. Here, each domain $\mathcal{D}_i$ represents a dataset $\{ (x_k^i, y_k^i)\}_{k=1}^{N_i}$ for each $i= 1, 2, \dots, S, T$, where $N_i$ is the number of instances in the domain $\mathcal{D}_i$, and $(x_k^i, y_k^i)$ denotes the input and output pair of the $k^{th}$ sample of the $i^{th}$ domain.

\subsection{Additive Modeling}
Fig. \ref{fig:model_figure} demonstrates the disentanglement of domain-specific and label features in our proposed approach. Given an image $x_i$, two different backbones are employed to extract features: $x_{label} = f_{\theta_{label}}(x_i)$ and $x_{domain} = f_{\theta_{domain}}(x_i)$. These backbones are trained with cross entropy loss, using their corresponding image and domain labels. From the cross entropy objective, it can be inferred that $x_{label}$ captures both domain variant and domain invariant features related to label, while $x_{domain}$ represents the domain-specific features. To disentangle the domain-specific information from the label features, an element-wise subtraction is performed:
\begin{equation}
    x_{dinv} = x_{label} - x_{domain}
\end{equation}
To optimize the domain invariant features $x_{dinv}$, adversarial domain discrimination loss, cross entropy loss and contrastive loss are utilized. This subtraction operation effectively prunes the domain-specific features while preserving the label information. Thus, $x_{dinv}$ contains the label information without the domain-specific features. Consequently, this design encourages the model to focus on extracting all the relevant information necessary to recognize an instance in $x_{label}$.

Optimizing $x_{dinv}$ with cross entropy ensures its effectiveness in performing the classification task. In this context, the domain-specific features serve as an "additional" guide, providing supplementary information to further improve classification performance. Moreover, in the process of optimizing $x_{label}$ and $x_{dinv}$, in-batch supervised contrastive loss \cite{khosla2020supervised} is employed to reduce the distance of the positive samples in the latent space. Such a metric-based loss is known to increase generalization \cite{kim2021selfreg, motiian2017unified} as it encourages compact decision boundaries.
\begin{equation}
    \mathcal{L}_{cont} = \sum_{i = 1}^N\frac{-1}{|P(i)|} \sum_{p \in P(i)}\log{\frac{\exp{(z_i \cdot z_p)}}{\sum_{a \in A(i)} \exp{(z_i \cdot z_a)}}} 
\end{equation}
\noindent where, $A(i)$ denotes the index set of all samples except the $i^{th}$, $P(i)$ denotes the index set of positive samples for $i$, and $|P(i)|$ is its cardinality. Each $z$ denotes the normalized latent feature vector for either $x_{label}$ or $x_{dinv}$.

\subsection{Remix Loss}
Exploiting the additive structure, the domain invariant features $x_{dinv}$ can be utilized and incorporated with another sample's domain features, $x_{domain}'$. This enables remixing data by combining samples from different domains but with the same labels.
\begin{equation}
    x_{remixed} = x_{dinv} + x_{domain}'
\end{equation}
Here, $x$ and $x'$ represent samples from different domains but with same labels. This method enables us to populate data by remixing in-batch samples during training. $x_{remixed}$ can then be used with the same classification module that maps $x_{label}$ to the logits. The addition of remixed samples further regularizes our model by enhancing its robustness to mixed samples from different domains. This can be seen as a form of data augmentation in the latent space, similar to the concept of mixup \cite{zhang2017mixup}. Notably, by leveraging weight sharing, increasing the complexity of the proposed model is prevented.

To compute the remix loss, the cross entropy loss is used as the classification objective:
\begin{equation}
    \mathcal{L}_{remix} = -\sum y_i \log(\hat{y}_i)
\end{equation}
\noindent where, $\hat{y} = f_{clf}(x_{remixed})$ represents the predicted logits obtained from the classification module. By incorporating the remix loss alongside the existing cross entropy loss, ADRMX learns more robust and discriminative representations and improve overall classification performance.

\subsection{Training Procedure}
The training procedure consists of optimizing two different losses in an alternating fashion, following the approach outlined in \cite{ganin2015unsupervised}. Similar to the generator-discriminator architecture in \cite{goodfellow2020generative}, the steps for the generator and discriminator losses are alternated.

In the generator loss, several losses are combined. These include the cross entropy losses related to the classification tasks, remix loss, contrastive loss, and discriminator loss. The discriminator loss, which detects the domain of $x_{dinv}$, is negated and scaled by a hyperparameter $\lambda$.

In the alternating step, the focus is on optimizing the discriminator loss, which consists solely of the discriminator's cross-entropy loss on domain classification. The generator and discriminator losses can be expressed as follows:

\begin{equation}
\mathcal{L}_{total\_gen} = \mathcal{L}_{CE} + \mathcal{L}_{remix} + \mathcal{L}_{cont} - \lambda \cdot \mathcal{L}_{disc}
\end{equation}
\begin{equation}
\mathcal{L}_{total\_disc} = \mathcal{L}_{disc}
\end{equation}
This alternating optimization procedure allows the generator to focus on improving the classification, contrastive, and remix objectives while taking into account the domain discrimination, guided by the discriminator loss. On the other hand, the discriminator, aims to correctly classify the domain of $x_{dinv}$ samples. By iteratively optimizing these losses, the model learns to disentangle domain-specific and domain-invariant features, incorporate remixing for data augmentation, and improve its overall performance in domain generalization tasks. During inference, class probabilities are obtained solely by utilizing the label encoder and label classifier.

\begin{figure*}[h]
  \centering
    \subfloat[]{\label{fig:fig2_a}\includegraphics[width=0.4\textwidth]{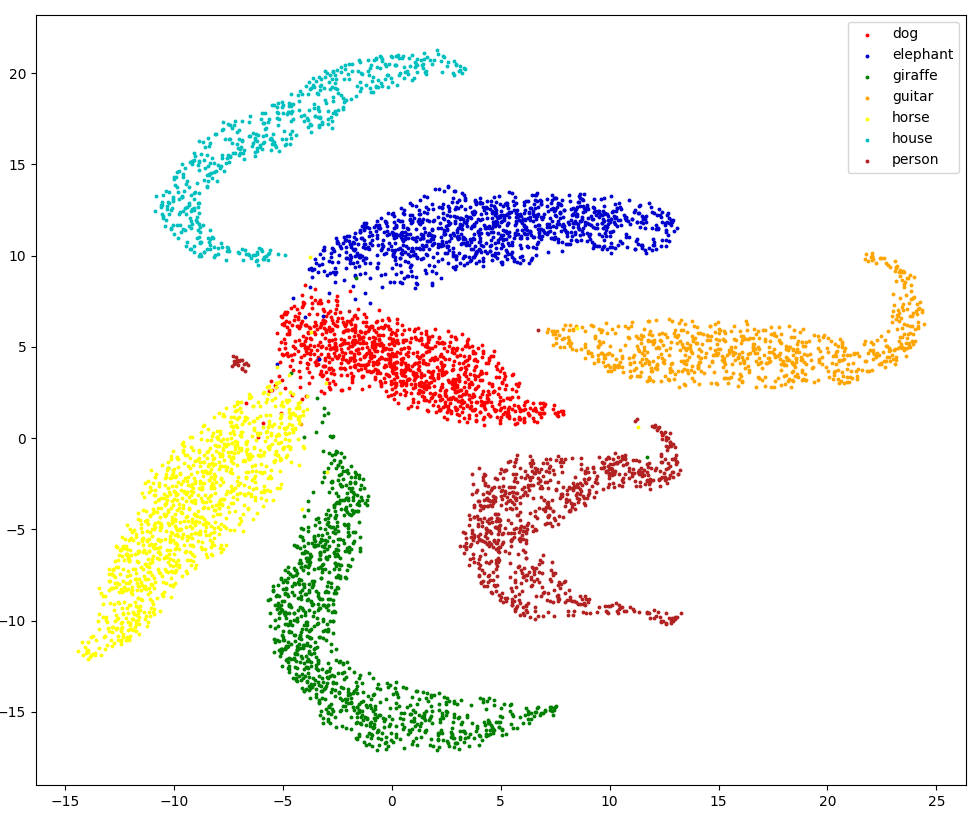}}
    \subfloat[]{\label{fig:fig2_b}\,\includegraphics[width=0.4\textwidth]{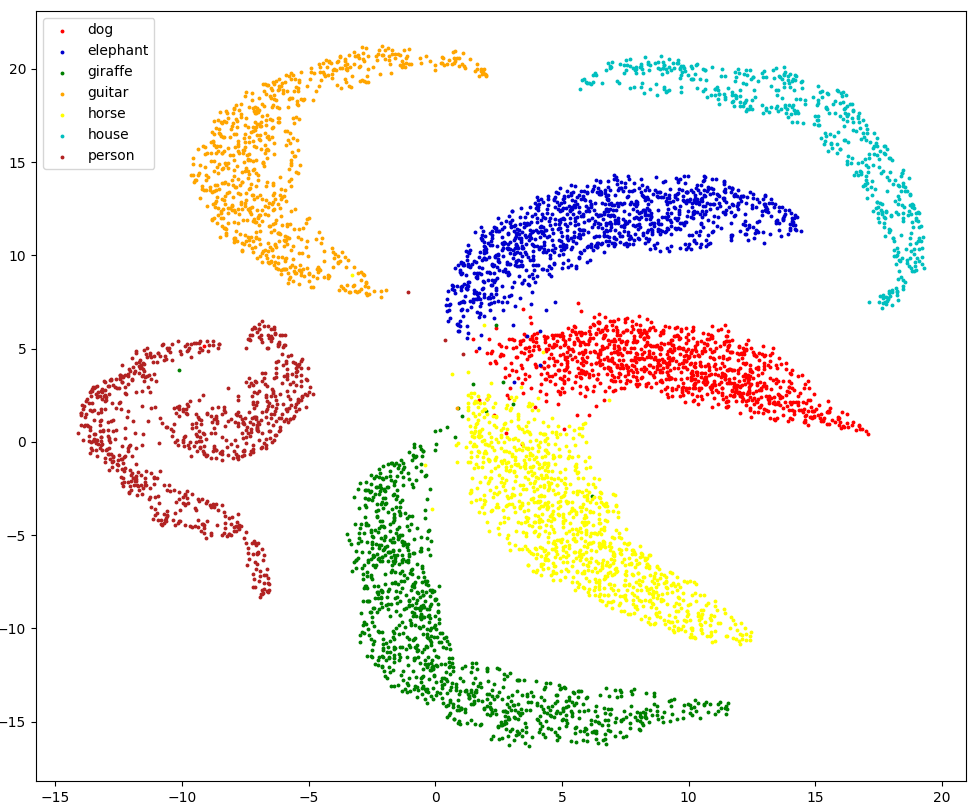}}\\
    \subfloat[]{\label{fig:fig2_c}\includegraphics[width=0.4\textwidth]{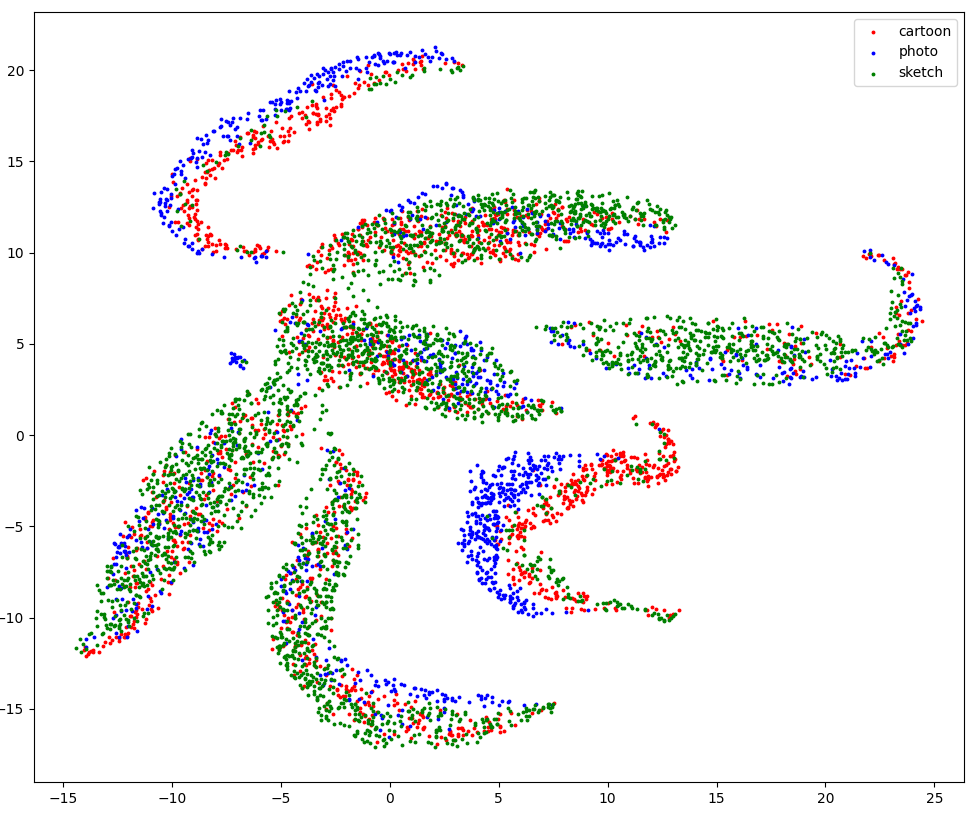}}
    \,\subfloat[]{\label{fig:fig2_d}\includegraphics[width=0.4\textwidth]{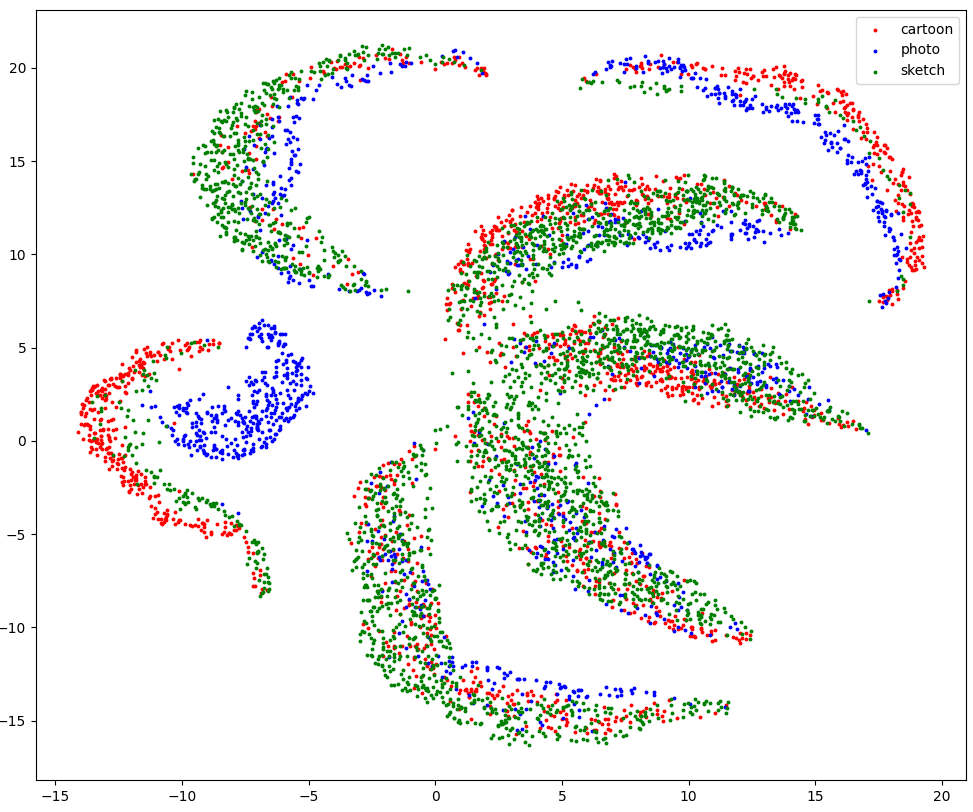}}
  
  \caption{UMAP visualization of the penultimate layer embeddings. The first row displays visualizations of domain-specific and domain-invariant features, with colors indicating class labels. The second row illustrates the same embeddings, now using a color map to represent domain labels. It can be observed that both features contain object information, while the domain-specific features potentially capture multimodalities across domains.}
    \label{fig:fig2}
\end{figure*}

\section{Experiments}
In this section, we provide implementation details of ADRMX, and present the experiments conducted using the \cite{gulrajani2020search} environment, consisting of $7$ datasets. Moreover, state-of-the-art comparison and an ablation study are performed to demonstrate the effectiveness of the model and its individual components.

\subsection{Implementation Details}
We use the DomainBed \cite{gulrajani2020search} environment, which provides a modular and easy-to-modify PyTorch \cite{paszke2019pytorch} codebase. Any proposed algorithm can be included in the environment by inheriting from the $Algorithm$ class and overriding $update$ and $predict$ methods. Our proposed algorithm is integrated by simply filling in the necessary components. Pretrained ResNet-50 \cite{he2016deep} with ImageNet \cite{deng2009imagenet} weights is used as a backbone architecture for both label and domain feature encoders. These backbones transform the images into a $2048$ dimensional latent space. To facilitate the domain discrimination, the adversarial learning technique is employed, as described in the Appendix A of \cite{ganin2015unsupervised}. This technique incorporates a GAN-like mechanism that replaces the gradient reversal layer with two different loss functions for the domain classifier \cite{goodfellow2020generative}. By alternating between these loss functions, positive and negative updates are performed. We optimize our network using the ADAM \cite{kingma2014adam} optimizer. Note that the hyperparameters for each dataset is determined using \cite{gulrajani2020search}'s random hyperparameter search, except for DomainNet \cite{peng2019moment}. The selection was based on the train domain validation set performance for each configuration. However, due to limited computational resources and the large search space, the number of hyperparameter configurations had to be restricted.

In the case of DomainNet, conducting an extensive hyperparameter search was infeasible due to the dataset's size. Therefore, we adopted the hyperparameters from TerraIncognita as a reasonable choice without performing a hyperparameter search. TerraIncognita was selected because it is the second largest dataset, providing a valuable baseline for comparison.

\begin{table*}[h]
\centering
\begin{adjustbox}{width=1\textwidth}
\begin{tabular}{*9l} 
    \toprule
    Model & CMNIST & RMNIST & VLCS & PACS & OfficeHome & TerraIncognita & DomainNet & Average \\ 
    \midrule
    ADRMX (ours) & 52.5 & 97.8 & 78.5 & 85.3 & 68.3 & 47.4 & \textbf{43.1} & \textbf{67.6} \\
    \midrule
    CORAL  \cite{sun2016deep} & 51.5 & 98.0 & \textbf{78.8} & 86.2 & \textbf{68.7} & 47.6 & 41.5 & 67.5 \\
    SagNet \cite{nam2021reducing} & 51.7 & 98.0 & 77.8 & \textbf{86.3} & 68.1 & \textbf{48.6} & 40.3 & 67.2 \\
    SelfReg \cite{kim2021selfreg} & 52.1 & 98.0 & 77.8 & 85.6 & 67.9 & 47.0 & 41.5 & 67.1 \\
    Mixup \cite{yan2020improve} & 52.1 & 98.0 & 77.4 & 84.6 & 68.1 & 47.9 & 39.2 & 66.7\\
    MLDG \cite{li2018learning} & 51.5 & 97.9 & 77.2 & 84.9 & 66.8 & 47.7 & 41.2 & 66.7 \\
    ERM \cite{vapnik1999overview} & 51.5 & 98.0 & 77.5 & 85.5 & 66.5 & 46.1 & 40.9 & 66.6 \\
    MTL \cite{blanchard2021domain} & 51.4 & 97.9 & 77.2 & 84.6 & 66.4 & 45.6 & 40.6 & 66.2 \\
    RSC \cite{huang2020self} & 51.7 & 97.6 & 77.1 & 85.2 & 65.5 & 46.6 & 38.9 & 66.1 \\
    ARM \cite{zhang2021adaptive} & \textbf{56.2} & \textbf{98.2} & 77.6 & 85.1 & 64.8 & 45.5 & 35.5 & 66.1 \\
    DANN \cite{ganin2016domain} & 51.5 & 97.8 & 78.6 & 83.6 & 65.9 & 46.7 & 38.3 & 66.1 \\
    VREx \cite{krueger2021out} & 51.8 & 97.9 & 78.3 & 84.9 & 66.4 & 46.4 & 33.6 & 65.6 \\
    CDANN \cite{li2018deep} & 51.7 & 97.9 & 77.5 & 82.6 & 65.8 & 45.8 & 38.3 & 65.6 \\
    IRM \cite{arjovsky2019invariant} & 52.0 & 97.7 & 78.5 & 83.5 & 64.3 & 47.6 & 33.9 & 65.4 \\
    GroupDRO \cite{sagawa2019distributionally} & 52.1 & 98.0 & 76.7 & 84.4 & 66.0 & 43.2 & 33.3 & 64.8 \\
    MMD \cite{li2018domain} & 51.5 & 97.9 & 77.5 & 84.6 & 66.3 & 42.2 & 23.4 & 63.3 \\
    \bottomrule
\end{tabular}
\end{adjustbox}
\vspace{1.5mm}
\caption{Comparisons with SOTAs on the DomainBed environment. Experiments are conducted based on train domain validation model selection.} 
\label{tables:1}
\end{table*}

\subsection{Experiments on DomainBed}
After carefully examining the model selection criteria proposed by \cite{gulrajani2020search}, we adopt train domain validation method for our experiments. It is efficient in two main ways: (1) it eliminates the need for performing cross domain testing, which significantly reduces the computation time; and (2) unlike oracle (test domain validation) selection method, it does not peek at the test set during performance evaluation, preserving the integrity and fairness of the evaluation process.

DomainBed benchmark includes a total of $7$ datasets.

\begin{itemize}
    \item \textbf{ColoredMNIST} is a synthetic dataset which builds on the MNIST handwritten digit classification dataset \cite{lecun1998mnist}. It has $3$ domains $\{+90\%, +80\%, -90\%\}$ with two labels, where the percentages indicate the degree of correlation between color and label. The dataset comprises $70,000$ images with a resolution of $2\times28\times28$ \cite{arjovsky2019invariant}.

    \item \textbf{RotatedMNIST} is constructed using MNIST as well. There are $6$ domains obtained with $15\%$ rotations ranging from $0$ to $90$ degrees. The dataset includes $10$ classes and consists of $70,000$ images with a resolution of $1\times28\times28$ \cite{ghifary2015domain}.
    
    \item \textbf{PACS} is a dataset which introduces a larger domain shift compared to the others, as it requires extracting higher semantic information to distinguish the same object from different domains. It consists of $9,991$ instances across $4$ domains $\{photo, art, cartoon, sketch\}$ and includes 7 classes. The images have a resolution of $3\times224\times224$ \cite{li2017deeper}.
    
    \item \textbf{VLCS} is built by merging $4$ datasets $\{$\textit{Caltech101, PASCAL VOC, LabelMe, SUN09}$\}$ each of which serves as a domain. The dataset contains a total of $10,729$ examples with a resolution of $3\times224\times224$ and $5$ classes \cite{fang2013unbiased}.
    
    \item \textbf{OfficeHome} has $15,588$ images across the domains $\{art, clipart, product, real\}$. The images in the dataset have a resolution of $3\times224\times224$ and belong to $65$ categories of everyday objects \cite{venkateswara2017deep}.
    
    \item \textbf{TerraIncognita} is a dataset consisting of wild animal photographs, it introduces $4$ domains based on the location where the images were captured $\{L100, L38, L43, L46\}$. It is the second largest dataset in DomainBed benchmark comprising $24,788$ images with a resolution of $3\times224\times224$ and $10$ different classes \cite{beery2018recognition}.
    
    \item \textbf{DomainNet} is the largest dataset in the benchmark containing $586,575$ instances from six domains $\{clipart, infograph, painting, quickdraw, real, sketch\}$. The dataset spans across $345$ distinct classes, and each image has a resolution of $3\times224\times224$ \cite{peng2019moment}.
\end{itemize}

DomainBed provides the performances of $14$ algorithms on the aforementioned datasets. For a fair comparison, the methods are compared under the same exact conditions, which include using the train domain validation model selection, limiting the number of hyperparameter configurations, using fixed backbone options and applying the same data augmentation techniques. For each hyperparameter configuration, the results of $3$ runs with different random initializations are averaged to report the final performance.

Table \ref{tables:1} shows that ADRMX achieved state-of-the-art performance. It outperformed the baseline ERM \cite{vapnik1999overview} and even surpassed the strongest work CORAL \cite{sun2016deep}, with an average accuracy of $67.6\%$. Comparing with the adversarial techniques SagNet \cite{nam2021reducing}, DANN \cite{ganin2016domain} and CDANN \cite{li2018deep}, ADRMX remarkably achieved improvements of $0.4\%$, $1.5\%$ and $2\%$ better than its compeers respectively. The instability of adversarial learning was commented on by SelfReg \cite{kim2021selfreg}, and this issue was addressed by reducing the learning rate and increasing the size of the discriminator network. Experiments demonstrated that the proposed additive disentanglement of domain and label features unraveled various possibilities to regularize training with different domains, such as remix loss. Due to the low learning rate and alternating updates - which effectively halve the iterations used for the generator's optimization - the number of epochs performed on the DomainNet \cite{peng2019moment} had to be increased as saturation was not reached by our model. The main reason for the increased iterations in DomainNet is that the size of the dataset, while other datasets did not require such extensions in training. In our view, this does not violate the fairness condition since the training domain validation model selection method focuses on the performance on the validation set. As shown in Table \ref{tables:2} even without the remix loss ADRMX outperforms CORAL \cite{sun2016deep} which is the strongest algorithm among $15$ evaluated, on the most challenging dataset, DomainNet. Overall, the empirical study supports our hypothesis that additive modeling can benefit from the domain-specific label-relevant information on top of the domain-invariant features.

\subsection{Ablation Study}
In this section we conduct an ablation study to assess the effectiveness of different components and design choices in our model. Specifically, the impacts of using contrastive loss, domain variant features' prediction, and remix loss are investigated. The results presented in Table \ref{tables:2} and \ref{tables:3} are based on averaging $3$ runs on the DomainNet \cite{peng2019moment} and PACS \cite{li2017deeper} datasets, respectively. Features from the penultimate layer are visualized on Figure \ref{fig:fig2} with respect to the domain label for both domain variant and domain invariant features on the PACS dataset using UMAP \cite{mcinnes2018umap}. The visualizations reveal the presence of subclusters within the same label, representing different modalities. This suggests that the model's representation is capable of capturing the multi-modal nature of the data, providing a relaxation over domain invariant feature extraction. It is also supported by the performance evaluation in Table \ref{tables:3} in which ADRMX performs better when domain variant features is used consistently for all domains, with an increase of $1.86\%$.

\begin{table}[h]
\centering
\begin{tabular}{*8l} \toprule
    Model & clip & info & paint & quick & real & sketch & Avg \\ 
    \midrule
    ADRMX & \textbf{60.8} & \textbf{20.9} & 48.6 & \textbf{14.1} & \textbf{61.8} & \textbf{52.4} & \textbf{43.1} \\
    ADRMX w/o remix loss & 59.6 & 20.6 & \textbf{50.3} & 12.5 & 61.5 & 50.3 & 42.4 \\
    CORAL \cite{sun2016deep} & 58.7 & \textbf{20.9} & 47.3 & 13.6 & 60.2 & 50.2 & 41.8 \\
\bottomrule
\end{tabular}
\vspace{1.5mm}
\caption{Ablation study on remix loss and comparison with state-of-the-art on DomainNet dataset.}
\label{tables:2}
\end{table}

As for the use of contrastive loss \cite{khosla2020supervised}, a performance jump of $1.21\%$ is observed. The incorporation of a distance/similarity-based loss encourages the model to learn compact decision boundaries, leading to the extraction of more robust features with across different domains. This finding aligns with previous work SelfReg \cite{kim2021selfreg}, which demonstrates the benefits of contrastive regularization in enhancing generalization performance. Furthermore, the impact of the remix loss on our model's performance is evaluated. Table \ref{tables:2} demonstrates that ADRMX achieves a significant performance jump of $0.7\%$ on the DomainNet dataset when trained with the remix loss. 

\begin{table}[h]
\centering
\begin{tabular}{*6l} \toprule
    Model & art & cartoon & photo & sketch & Avg \\ 
    \midrule
    ADRMX (original) & \textbf{87.69} & \textbf{80.55} & \textbf{97.74} & \textbf{77.53} & \textbf{85.87} \\
    ADRMX w/domain invariant & 85.48 & 76.79 & 97.64 & 76.1 & 84.01 \\
    ADRMX w/o contrastive & 87.05 & 78.56 & 97.03 & 76 & 84.66 \\
\bottomrule
\end{tabular}
\vspace{1.5mm}
\caption{Ablation study on using contrastive loss and domain invariant features on PACS dataset.}
\label{tables:3}
\end{table}

Inspired by the mixup technique \cite{zhang2017mixup}, we adopt an additive approach that separates the domain-specific component from the overall feature vector. This design enables the removal and addition of domain information, facilitating the augmentation and diversification of the training data.

\section{Conclusion}
In this paper, we presented ADRMX, a domain generalization approach that disentangles the domain variant and domain invariant features in an additive fashion. Unlike previous methods, domain variant features are effectively utilized alongside the domain invariant ones. Moreover, a latent space data augmentation technique is introduced to further enhance the generalization capabilities of our model. Through comprehensive experiments on the DomainBed benchmark, ADRMX demonstrated outstanding performance compared to 14 other models across 7 diverse datasets under fair conditions. It achieved state-of-the-art results, reaffirming its effectiveness and robustness under domain shift scenarios. By effectively capturing the contextual characteristics of objects and leveraging the additive modeling approach, ADRMX showcases its potential for addressing the challenges posed by domain shift in real-world applications.

\bibliographystyle{unsrt}  
\bibliography{references}

\end{document}